\documentclass[10pt]{article}

   \usepackage[pdftex]{graphicx}
   \DeclareGraphicsExtensions{.pdf,.jpeg,.png}

\usepackage{vmargin}
\usepackage{amsmath}
\newcommand{\g}[1]{\ensuremath{\mathbf{#1}}}

\usepackage{algorithm}
\usepackage{mathtools}

\PassOptionsToPackage{noend}{algpseudocode}
\usepackage{algpseudocode}
\makeatletter

\usepackage[colorinlistoftodos]{todonotes}

\usepackage{hyperref}
\newcommand{\footremember}[2]{%
    \footnote{#2}
    \newcounter{#1}
    \setcounter{#1}{\value{footnote}}%
}
\newcommand{\footrecall}[1]{%
    \footnotemark[\value{#1}]%
} 

\begin{document}
%
\title{Two-pixel polarimetric camera by compressive sensing}%
%
%

\author{Julien Fade\footremember{foton}{Institut FOTON, University of Rennes~1, CNRS, Campus de Beaulieu, Rennes, France},
        Est\'eban Perrotin\footrecall{foton},
        and~J\'er\^ome Bobin\footnote{CEA, IRFU, Service d'Astrophysique-SEDI, Gif-sur-Yvette, France}
}

%
%



\maketitle

\begin{abstract}
  We propose an original concept of compressive sensing (CS)
  polarimetric imaging based on a digital micro-mirror (DMD) array and
  two single-pixel detectors. The polarimetric sensitivity of the
  proposed setup is due to an experimental imperfection of reflecting
  mirrors which is exploited here to form an original reconstruction
  problem, including a CS problem and a source separation task. We
  show that a two-step approach tackling each problem successively is
  outperformed by a dedicated combined reconstruction method, which is
  explicited in this article and preferably implemented through a
  reweighted FISTA algorithm. The combined reconstruction approach is
  then further improved by including physical constraints specific to
  the polarimetric imaging context considered, which are implemented
  in an original constrained GFB algorithm. Numerical simulations
  demonstrate the efficiency of the 2-pixel CS polarimetric imaging
  setup to retrieve polarimetric contrast data with significant
  compression rate and good reconstruction quality. The influence of
  experimental imperfections of the DMD are also analyzed through
  numerical simulations, and 2D polarimetric imaging reconstruction
  results are finally presented.
\end{abstract}

\section{Introduction}

In various application domains such as biomedical
diagnosis, defence or remote sensing, standard intensity imaging
techniques sometimes fail to reveal relevant contrasts or to gather
sufficient information. In these domains, polarimetric approaches have
proved efficient to enhance the estimation or detection capabilities
of the imaging systems \cite{gro99,bou05,ana08,
  bre99,alo09,fad14,sni14}. Mostly often, the polarimetric information
is provided by a scalar polarization contrast image, which offers
complementary constrast information with respect to the conventional intensity
image. This polarization contrast image usually corresponds to the map
of the degree of polarization (DOP) of the light backscattered at each
location of the scene. Four polarimetric measurements are
theoretically needed to determine such a DOP image
\cite{bro98}. However, to reduce both cost and acquisition time,
simplified 2-channel imaging modalities are usually preferred in
active polarimetric imaging, such as in Orthogonal States Contrast
(OSC) imaging. This approach consists of computing a contrast map
$\mathrm{OSC}=(\g{x}_S-\g{x}_P)/(\g{x}_S-\g{x}_P)$ between two
polarimetric images ($\g{x}_S$ and $\g{x}_P$) of the scene, acquired
through a linear polarizer oriented along two orthogonal directions,
denoted $S$ and $P$ throughout the article where $S$ denotes the
polarization direction of the illumination source. Owing to its
instrumental simplicity, which can be further improved using
voltage-controlled electro-optics devices \cite{gup02,ann12} or appropriate
optical design \cite{ben09}, and due to the fact that OSC provides a
good estimate of the DOP (under the assumption that the imaged objects
are purely depolarizing \cite{bre99,gou04b}), OSC imaging is today
widely used in many applications \cite{gou04b}.

On the other hand, compressive sensing (CS) and CS-derived original
imaging concepts such as the single-pixel camera (SPC) have attracted
much attention these past years \cite{bar08,cha08b}. With this
approach, the measurement process relies on the spatial sampling of
the image of interest with a Digital Micromirror Device (DMD), and on
numerical reconstruction of the image from intensity measurements on a
single photodetector for different sampling patterns on the DMD,
allowing a compressed version of the image to be recovered from the
photocurrent signal acquired. More recently, the concept of SPC has
been applied to a number of domains including, among others,
multi/hyperspectral imaging \cite{wag08,ase10,stu12,aug13}, THz
imaging \cite{cha08b}, or random media-assisted CS
\cite{liu13}. However, despite the swarming interest in CS, only few
attempts were reported so far to perform polarimetric CS imaging
\cite{ase10,dur12,sol13, wel15,fu15}. The imaging setups proposed in
these references are all directly based on the SPC concept, where
polarimetric sensitivity was simply gained by detecting the optical
signals through appropriate polarization analyzing devices during
sequential acquisitions, or with a unique acquisition on several
detectors after appropriate beam splitting of the light reflected by
the DMD. In these references, the reconstruction process consisted of
solving as many CS reconstruction problems as polarimetric channels
were considered (2 or 4). More precisely, in \cite{dur12,sol13}, the
SPC scheme was readily improved by adding a rotating polarizer in
front of the detector. These techniques can operate as polarimetric
imaging systems only at the expense of a two-fold (respectively
four-fold) increase in the measurement time, while at the same time
suffering from the loss in intensity due to the use of a polarization
analyzer. In references \cite{wel15,fu15}, the measurement time was
limited, but the complexity of the imaging system was largely
increased, to the expense of important losses in the imaging setup.

In this article, we revisit the problem of 2-channel polarimetric CS
by proposing an original polarimetric imaging architecture using two
single-pixel detectors. The proposed setup is still inspired from the
initial concept of SPC, but does not require any polarization
analyzing element as it relies on imperfections of the DMD
itself. Contrarily to previous attempts in polarimetric CS, the
polarimetric information is obtained through a single temporal data
acquisition on the two photodetectors, and the polarimetric channels
are recovered simultaneously from a single reconstruction step. It
also offers in principle the best detectivity tradeoff, as all the
light directed towards the DMD is involved in the imaging process
without passing through any polarization analysis component.

This article is organized as follows: in Section \ref{principle}, the
principle and the optical setup proposed to achieve CS polarimetric
imaging are detailed. Then in Section \ref{algos}, various algorithms
are presented to tackle this original CS inverse problem, along with
several possible algorithmic optimizations such as reweighted
approaches or constrained algorithms that can be implemented to
improve the reconstruction results. The performance of the algorithms
and of the 2-pixel polarimetric CS imaging approach are finally discussed
in Section \ref{results} through numerical simulations on 1D and 2D
test signals. In this section, the influence of experimental
imperfections on the reconstruction quality is also analyzed, before
conclusions and perspectives of the article are provided in Section
\ref{conclu}.

\section{Principle of 2-pixel CS polarimetric imaging}\label{principle}
  
\subsection{2-pixel CS polarimetric imaging setup}\label{setup}
\begin{figure}[t!]
\begin{center}
\includegraphics[width=8.5cm]{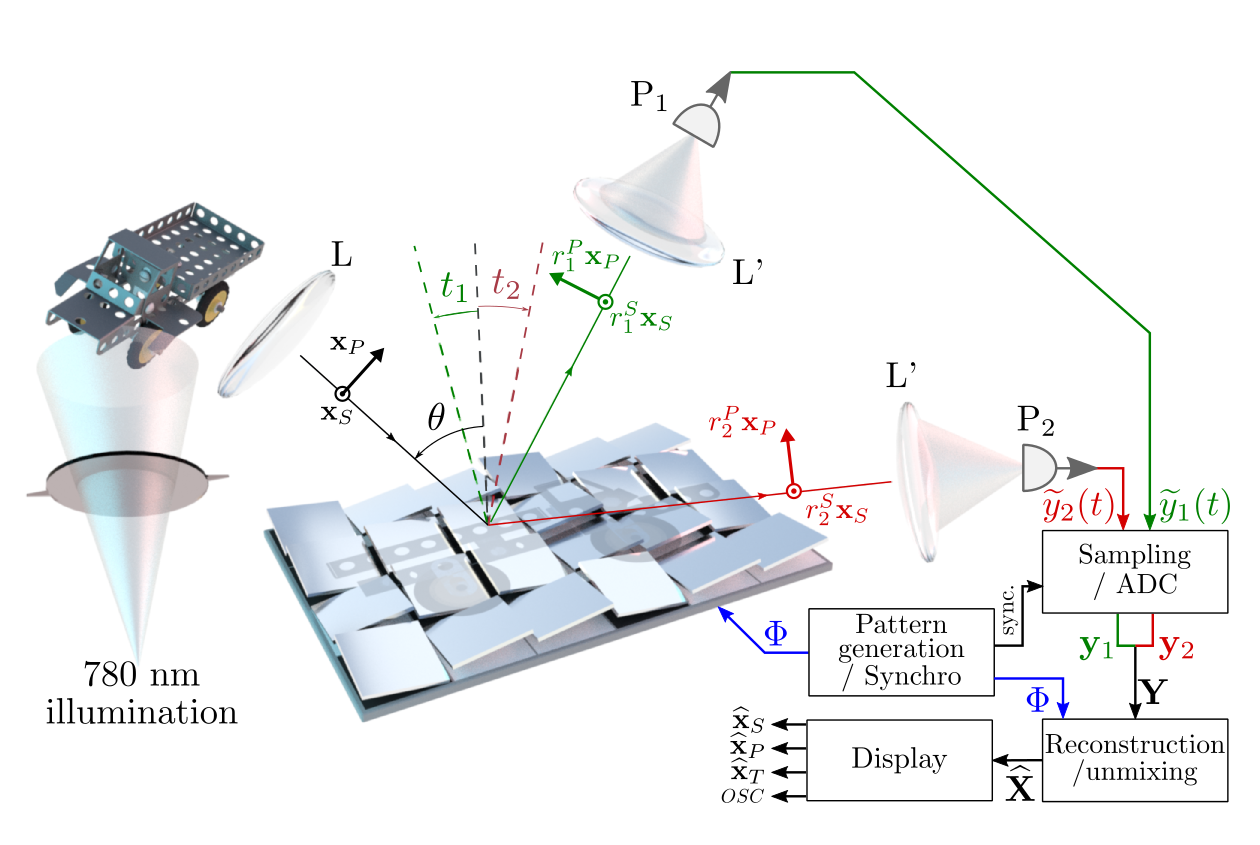}
\end{center}
\caption{Sketch of the 2-pixel CS polarimetric imaging setup
  proposed. It is inspired from the concept of SPC where the
  image is first spatially sampled by a DMD which reflects light in
  two directions, and where the total photon flux is detected on a
  single photodetector in each direction. \label{fig:setup}}
\end{figure}

The polarimetric CS imaging approach proposed relies on the SPC
imaging architecture. As will be demonstrated, it makes it possible to
perform standard intensity and polarimetric contrast imaging using CS,
without requiring any polarization analysis component. The
corresponding experimental setup is sketched in
Fig.~\ref{fig:setup}. In the context of active imaging, we assume that
the scene or object of interest is enlightened by an horizontally
polarized light source. An image of the scene is formed through a lens
$L$ onto the surface of the DMD, which spatially samples it by
applying a controlled binary pattern on the micromirrors. The
detection setup is strictly similar to the SPC architecture, with a
first photodetector ($P_1$) used to detect light reflected in the
first reflection direction through a lens ($L'$). However, the light
reflected in the second direction of the tilted mirrors is directed
towards a second photodetector ($P_2$) via a lens ($L'$), instead of
being discarded as in the original SPC scheme. By applying a series of
different patterns on the DMD, the detection of the total light
intensity reflected in directions 1 and 2 by photodetectors $P_1$ and
$P_2$ provides two temporal signals that are sampled and digitized on
a Analog to Digital Converter (ADC), before they can be used for
numerical inversion of the intensity and polarimetric contrast images.

Throughout this article, we will denote the total intensity image of
the scene by a single dimensional row vector
$\mathbf{x}_T=\{x_{T_i}\}_{i=1,\ldots,N}$ of length $N$. We assume
that the total intensity image can be written as the sum of two
polarimetric components, i.e.,
$\mathbf{x}_T=\g{x}_S+\g{x}_P$. Subscripts $S$ and $P$ denote two
orthogonal linear polarization directions w.r.t. the orientation of
the DMD surface and to the direction of linear polarization of the
illuminating beam ($S$), as sketched in Fig.~\ref{fig:setup}. We
assume that these two components are compressible in the same sparse
representation $\{\Psi_k\}_{k=1,\ldots,N^\prime}$, i.e., they can be written as
$\g{x}_{S,P}=\g{s}_{S,P}\,\boldsymbol{\Psi}$, where
$\boldsymbol{\Psi}=[\Psi_{1} , \ldots, \Psi_{N^\prime}]$ is a $N^\prime\times N$
matrix with $N^\prime\geq N$, and $\g{s}_{S,P}$ are $N^\prime \times 1$ column vectors containing
the expansion coefficients. In the compressed sensing framework,
compressibility means that most of these expansion coefficients have a
small amplitude. Only the few large-amplitude coefficients code for
the salient information of the polarimetric signals. This assumption
generally refers to the approximately sparse signal model in the CS
literature \cite{EldarBook12}.

Upon reflection on the DMD surface, the polarimetric components of the
original image $\g{x}_T$ are altered by the Fresnel's reflection
coefficients (in intensity) of the mirror, depending on the reflection
angle. Let us denote by $r_i^S$ and $r_i^P$ the Fresnel's reflection
coefficients (in intensity) of the mirror for each tilt direction
$i=1,2$ and respectively for the $S$ and $P$ polarimetric components
of the image formed on the DMD surface. Neglecting absorption effects,
these coefficients are real and verify $r_{1,2}^{S,P}\in[0,1]$. With
such notations, the images respectively reflected towards detectors
$P_1$ and $P_2$ read
$\widetilde{\g{y}}_1^{\circ}=r_1^S \g{x}_S + r_1^P \g{x}_P$ and
$\widetilde{\g{y}}_2^{\circ}=r_2^S \g{x}_S + r_2^P \g{x}_P$, which can
be rewritten in a compact form as
\begin{equation}
\widetilde{\g{Y}}^\circ=\left[\begin{array}{c}  \widetilde{\g{y}}_1^\circ \\  \widetilde{\g{y}}_2^\circ \end{array}\right]=\left[ \begin{array}{cc} r_1^S & r_1^P \\ r_2^S  &r_2^P \end{array}\right]\left[\begin{array}{c}  \g{x}_S \\  \g{x}_P \end{array}\right]=\widetilde{\mathrm{\mathbf{A}}} \g{X}.
\end{equation}

When a given pattern indexed by $k$ is applied on the DMD to spatially
sample the image, the total intensity reflected towards direction 1
and integrated on detector $P_1$ can be denoted by
$\widetilde{y}_1^{(k)}=\widetilde{\g{y}}_1^{\circ} \varphi^{(k)} $, where $\varphi^{(k)}$ is a
binary valued $N$-dimensional column vector encoding the set of
orientations of the individual mirrors (DMD pattern).  Similarly,
$\widetilde{y}_2^{(k)}=\widetilde{\g{y}}_2^{\circ} \overline{\varphi}^{(k)}$, where
$\overline{\varphi}^{(k)}=\mathbf{1}_{N}-\varphi^{(k)}$ is the complement of
vector $\varphi^{(k)}$. Then, when $M$ measurements are accumulated with
various sets of pseudo-random configurations of the DMD, the detected
intensities $\{\widetilde{y}_i^{(1)} \ldots \widetilde{y}_i^{(M)}\}_{i=1,2}$ organized in a
$2\times M$ measurement matrix read
\begin{equation}
\widetilde{\g{Y}}=\left[\begin{array}{c} \widetilde{\g{y}}_1^\circ \boldsymbol{\varphi} \\\widetilde{\g{y}}_2^\circ \overline{\boldsymbol{\varphi}}\end{array} \right],
\end{equation}
with sensing matrix $\boldsymbol{\varphi}=[\varphi^{(1)} \ldots \varphi^{(M)}]$ and $\overline{\boldsymbol{\varphi}}=[\overline{\varphi}^{(1)} \ldots\overline{\varphi}^{(M)} ]$ .
    
Under the above assumptions, we will show that such a simple setup
suffices to retrieve a compressive measurement of the polarimetric
components $\g{x}_S$ and $\g{x}_P$, and thus of the intensity image
$\mathbf{x}_T=\g{x}_S+\g{x}_P$, and OSC image
$\mathrm{OSC}= (\g{x_S}-\g{x_P})/\mathbf{x}_T$. It is worth noting
here that all the available light incoming from the scene is detected,
thereby offering optimal energy balance, and that no polarimetric
optical component has been inserted in the setup. Indeed, to achieve
polarimetric sensitivity, this CS imaging setup relies on the slight
variation of the Fresnel coefficients for the $S$ and $P$-polarized
components of light as a function of the angle of incidence of the
incoming light beam. As sketched in Fig. \ref{fig:setup}, the angle of
incidence on the mirror is denoted by $\theta_1=\theta-t_1$
(respectively $\theta_2=\theta-t_2$) for tilt direction $1$
(respectively direction $2$), where $\theta$ is the angle of incidence
with respect to the DMD surface, and where $t_1$ and $t_2$ denote the
tilt angle of the mirrors (typically $t_1=12^\circ$ and
$t_2=-12^\circ$ on most DMDs \cite{specs}).  This dependency with
$\theta$ is illustrated in Fig. \ref{fig:CN}.a, where we plotted the
evolution of the four reflection coefficients $r_{1,2}^{S,P}$ at
wavelength 780~nm for aluminum mirrors when $\theta$ varies from
$17^\circ$ to $65^\circ$. The influence of the value of these
coefficients on the CS reconstruction problem is discussed below in
Section \ref{sec:expcond} and their calculation is recalled in
Appendix \ref{anex:1}. At this level, it is interesting to note that
polarization vision mechanisms in some animal species rely on such
polarization sensitivity of the Fresnel reflection or refraction
coefficients\cite{fla98}.

\subsection{Description of the CS inverse problem}

With the above notations, we will now describe this imaging process as
a CS inverse problem. We first rewrite the binary sensing matrix
$\boldsymbol{\varphi}$ as
$\boldsymbol{\varphi}= (\mathbf{1}_{N,M} +\boldsymbol{\Phi})/2$, and
consequently,
$\overline{\boldsymbol{\varphi}}=(\mathbf{1}_{N,M}
-\boldsymbol{\Phi})/2$, where the elemets of $\boldsymbol{\Phi}$ take
on $-1$ and $+1$ values. We impose that, for all DMD configurations,
50 $\% $ of the micromirrors are oriented towards direction 1 such
that $\sum_{i=1}^N \Phi_i^{(k)}=0$. In such a case, the measurement
matrix can then be rewritten as
$\widetilde{\g{Y}}=\overline{\g{Y}} +\g{Y}$, with
$\overline{\g{Y}} =\widetilde{\g{Y}}^\circ \mathbf{1}_{N,M}/2$, and
\begin{equation}\label{eq:Y}
  \g{Y}=\frac{1}{2}\left[\begin{array}{c} \widetilde{\g{y}}_1^\circ \boldsymbol{\Phi}\\ - \widetilde{\g{y}}_2^\circ \boldsymbol{\Phi}\end{array}\right]=\frac{\g{Q}\widetilde{\g{Y}}^\circ\boldsymbol{\Phi}}{2}, \quad\text{with}\quad \g{Q}=\left[\begin{array}{cc}1&0\\0&-1\end{array}\right].
\end{equation}

For the sake of simplicity, we will assume in the following that the
constant term $\overline{\g{Y}}$ can be easily estimated and
subtracted out from the measured data, e.g. by averaging the
acquisitions over all $M$ DMD pattern realizations considered. In this
case, the CS problem that must be solved is given in Eq.~(\ref{eq:Y})
and reads
$\g{Y}=\g{Q}\widetilde{\g{Y}}^\circ\boldsymbol{\Phi}/2=\g{Q}\widetilde{\g{A}}\g{X}\boldsymbol{\Phi}/2$. For
the sake of clarity, and taking into account an additive noise
contribution $\mathbf{b}$ on the detected intensities, we propose to
rewrite it with simplified notations as
\begin{equation}\label{CSpb}
  \g{Y}=\g{A}\g{X}\boldsymbol{\Phi}+\mathbf{b}, \quad\text{with}\quad \g{A}=\frac{\g{Q}\widetilde{\g{A}}}{2}=\frac{1}{2}\left[\begin{array}{cc}r_1^S&r_1^P\\-r_2^S&-r_2^P\end{array}\right].
\end{equation}
We also introduce $\g{Y}^\circ=\g{A}\g{X}$, such that $\g{Y}=\g{Y}^\circ\boldsymbol{\Phi}$.

As a result, the $N$-dimensional polarimetric components of the image
contained in $\g{X}$ can be in principle recovered from a number of
measurements $M\ll N$ provided the problem described in the above
equation can be solved. Contrarily to most CS inverse problems that
have been considered so far, we are facing an additional difficulty in
this particular situation, as the signals to recover are strongly
mixed in the measurement process via the mixing matrix
$\mathrm{\mathbf{A}}$. Indeed, as illustrated in Fig.~\ref{fig:CN}.a,
the reflection coefficients on metals are usually quite similar for polarization
directions $S$ and $P$, causing a strong crosstalk between the two
components of interest. As a consequence, the signals detected at
photodetectors $P_1$ and $P_2$ are almost perfectly anticorrelated,
the polarimetric information lying in the tiny discrepancies between
these two signals. This is illustrated in Fig.~\ref{fig:signals},
where simulated intensity signals are plotted. As will be shown in
Section~\ref{algos}, several approaches can be used to tackle this
unmixing/CS reconstruction problem, either by considering the two
problems independently, or by solving them simultaneously in the
recovery process.

\subsection{Experimental parameters and imperfections}\label{sec:expcond}

\begin{figure}[t]
\begin{center}
\includegraphics[width=8.5cm]{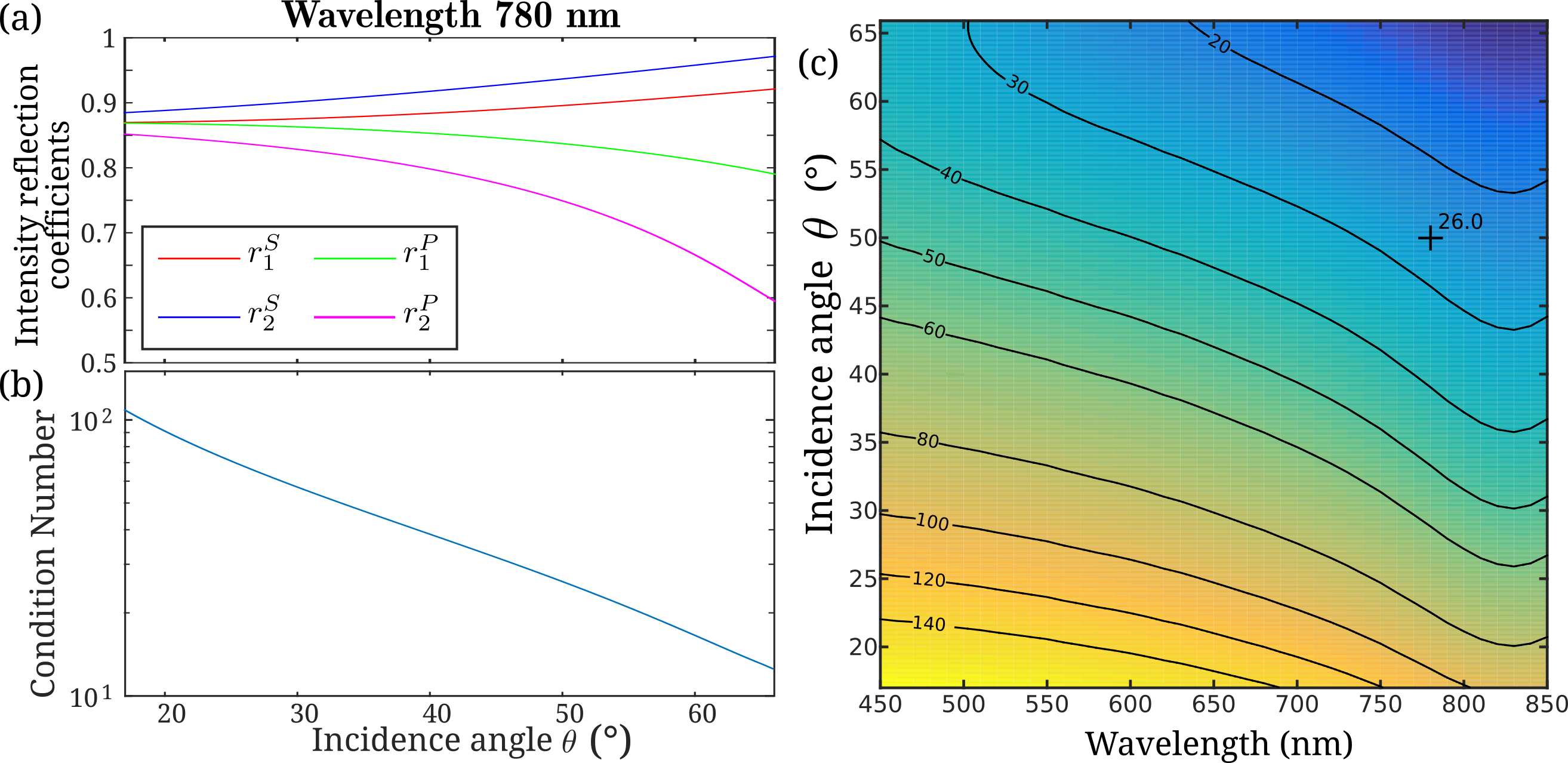}
\end{center}
\caption{(a) Evolution of the reflection coefficients in intensity for
  two tilt directions and two polarimetric components $S$ and $P$, as
  a function of incidence angle $\theta$ on the DMD surface at
  wavelength 780 nm. (b) Evolution of the condition number
  $\kappa(\mathrm{\mathbf{A}})$ of matrix $\mathrm{\mathbf{A}}$ as a
  function of $\theta$ at 780 nm. (c) Contour plot of
  $\kappa(\mathrm{\mathbf{A}})$ as a function of $\theta$ and
  wavelength. The black cross indicates the physical situation
  addressed in the numerical experiments.\label{fig:CN}}
\end{figure} 

Before we detail the reconstruction algorithms used to achieve
polarimetric CS imaging with the 2-pixel camera setup proposed, let us
analyze the possible influence of some experimental parameters on the
reconstruction quality, and how these parameters can be optimized.

Obviously, an important parameter that will control the difficulty of
the unmixing problem is the mixing matrix $\mathrm{\mathbf{A}}$, which
depends on the wavelength and bandwidth of the illuminating source, on
the incidence angles $\theta_1$ and $\theta_2$ on the two tilt
positions, and on the optical coating of the reflective surface of the
micromirrors. Sticking to the specifications of commercially available
DMDs \cite{specs}, we have simulated the values of the $r_{1,2}^{S,P}$ coefficients for
aluminum-coated metallic micromirrors with tilt angles $\pm 12^\circ$,
for varying wavelengths over the spectral bandwidth of commercially
available DMDs (450-850 nm), and for varying incidence angle
$\theta$. The expression of the Fresnel's intensity reflection
coefficients on metals is recalled in Appendix~\ref{anex:1}. The
evolution of parameters $r_{1,2}^{S,P}$ with $\theta$ is plotted in
Fig.~\ref{fig:CN}.a for a wavelength of 780 nm, showing that the four
reflection coefficients considered do not differ much (typically, less
that 15$\%$ difference for reasonable incidence angles). The unmixing
step in the reconstruction problem consisting basically of an
``inversion'' of matrix $\mathrm{\mathbf{A}}$, the condition number
$\kappa(\mathrm{\mathbf{A}})=\|\mathrm{\mathbf{A}}^{-1}\|_2 \cdot
\|\mathrm{\mathbf{A}}\|_2$ naturally gives an indication about the
difficulty of such inversion procedure. We have thus plotted in
Fig.~\ref{fig:CN}.b the evolution of $\kappa(\mathrm{\mathbf{A}})$ as
a function of $\theta$, which confirms that high incidence angles may
be favored. We further analyzed the evolution of
$\kappa(\mathrm{\mathbf{A}})$ with $\theta$ and with the illumination
wavelength. The contour plot in Fig.~\ref{fig:CN}.c shows that the
condition number can vary by a factor of 5 across the range of
wavelength and incidence angle considered, higher wavelengths being
best adapted to maximize the polarimetric sensitivity of the
reflection on aluminum mirrors. As a result, owing to the
availability of laser and LED sources at such wavelength, and to keep
reasonable incidence angles, we will consider throughout the remainder
of this article the situation of a monochromatic illumination at 780
nm, with incidence angle $\theta=50^\circ$ (black cross in
Fig.~\ref{fig:CN}), yielding a reasonably low condition number of
$\kappa(\mathrm{\mathbf{A}})=26$.

It can be noted here that dielectric coated micromirrors could offer
stronger polarization dependence of the reflection coefficients, but
to the expense of totally revisiting the fabrication process of
DMDs. For the sake of simplicity, we will neglect in this article the
influence of the anti-reflection coated cover slit that protects the
DMD surface \cite{specs}. We also neglect the dispersion of the values
of coefficients $r_{1,2}^{S,P}$ with the source spectral bandwidth and
with the slight variation of incidence angle when the image of the
scene is formed on the DMD surface. All these possible sources of
imperfections can be neglected here to study the principle of 2-pixel
polarimetric camera, but may be addressed in further work to achieve
the experimental validation of this CS imaging scheme.

Nevertheless, we shall analyze in our simulations how a bias in the
estimated incidence angle $\theta$ can have an impact on
reconstruction quality. Indeed, it is quite obvious that a bias on
$\theta$ will lead to using an incorrect mixing matrix in the
inversion process, hence hindering the recovery of a
satisfactory polarization contrast image. Moreover, we also consider
the influence of possible individual random errors in the tilt angles
of each micromirror of the DMD. Indeed, the typical individual tilt
error is about $\pm 1^\circ$ to $\pm 2^\circ$ according to standard
DMD's specifications \cite{specs}, thus its consequences on the
reconstruction process may not be negligible. If global angle bias on
$\theta$ and individual tilt errors can affect the reconstruction,
their influence on image quality is likely to be very
different. Global bias on $\theta$ can be basically treated as a
calibration error, whereas individual tilt errors would rather behave
as additional random noise on the inversion process.

Lastly, we will consider that the only source of noise is the
photodetectors electronic noise, and we assume statistical
independence between the noise at photodetectors $P_1$ and $P_2$, and
between noise realizations as the DMD patterns are varied. In the
above inverse problem, noise vector $\g{b}$ can thus be modeled by a
centered Gaussian distribution of variance $\sigma^2$, i.e.,
$\forall i\in \{1,2\},\forall j\in[1,M], \g{b}_{ij} \sim {\cal N}(0,\sigma^2)$.

\section{Reconstruction algorithms}\label{algos}

In this section, we describe different strategies to reconstruct the polarimetric
data $\g{X}$ from the compressed measurements $\g{Y} =
\g{A} \g{X}\boldsymbol{\Phi} + \mathbf{b}$. In the framework of
compressed sensing, the signal recovery problem is generally described
as a ``large $p$, small $n$'' problem, where the number of unknown,
i.e. the number of samples in the polarimetric components, can
be much larger than the number of measurements in $\g{Y}$. Hence,
it is essential to enforce additional constraints on the signal to be
recovered, which eventually boils down to solving a
minimization problem of the form
\begin{equation}
\label{eq:recovery1}
\widehat{\g{X}} = \mbox{Argmin}_{\g{X}} \quad \mathcal{P}\left(\g{X}
\right) + {\frac{1}{2} \left\| \g{Y} - \g{A} \g{X}\boldsymbol{\Phi} \right \|_F^2},
\end{equation}
where the first term is a penalization term that favors solutions with
certain desired properties, and the second term is a data fidelity
term that measures the discrepancy between the data $\g{Y}$ and the
model $\g{A} \g{X}\boldsymbol{\Phi}$. The Frobenius norm is defined as
$\left\| \g{Y} \right \|_F^2 =
\mbox{Trace } \left( \g{Y} \g{Y}^T\right)$.\\
In the context of CS \cite{CandesReview2,Candes:cs1}, it is customary
to enforce the sparsity of the unknown variable $\g{X}$ in some signal
representation $\boldsymbol{\Psi}$ that is chosen a priori. In the
following sections, we describe and compare several strategies that
are precisely dedicated to solve the two-pixel polarimetric compressed
sensing recovery problem.

\subsection{2-stage reconstruction approach}

The recovery problem in Eq.~\eqref{eq:recovery1} can described as the
combination of two classical inverse problems: a compressed sensing
problem and a source separation problem. A first straightforward
approach consists in tackling alternately both problems. Then,
recovering the polarimetric components can be performed with the
following 2-stage approach.

\begin{itemize}
\item{\bf Compressed sensing: } denoting the
  non-compressed mixed polarimetric components by
  $\g{Y}^\circ = \g{A} \g{X}$, the actual measurements $\g{Y}$ can be
  defined as $\g{Y} = \g{Y}^\circ \boldsymbol{\Phi} $. Recovering
  $\g{Y}^\circ$ then
  boils down to a standard compressed sensing recovery problem. This step is customarily solved by finding the minimum of the problem\\
\begin{equation}
\label{eq:recovery2}
\widehat{\g{Y}^\circ} = \mbox{Argmin}_{\g{Y}^\circ} \quad \left \|
  \boldsymbol{\Lambda} \odot\left( \g{Y}^\circ
 \boldsymbol{\Psi}^T \right)\right \|_{\ell_1} + {\frac{1}{2} \left\| \g{Y} - \g{Y}^\circ\boldsymbol{\Phi}  \right \|_F^2},
\end{equation}
where the $1$-norm $\|\, . \, \|_{\ell_1}$ enforces the sparsity of
$\g{Y}^\circ$ in $\boldsymbol{\Psi}$ and $\boldsymbol{\Lambda}$ stands
for the regularization parameters, which is composed of strictly positive entries (see
section \ref{sec:param_tuning} for details about the parameters' selection). This optimization problem is solved
using the reweighting FISTA algorithm. This algorithm
is referred to as Algorithm \ref{alg_rFISTA} and is detailed in Appendix \ref{anex:2}.\\

\item{\bf Source separation: } once the mixed polarimetric
   components $\g{Y}^\circ$ are estimated, retrieving $\g{X}$ from  $\g{Y}^\circ = \g{A} \g{X}$ is equivalent to a
  source separation or unmixing problem, which can be tackled by
  minimizing the Euclidean distance between $\g{Y}^\circ$ and the
  model $ \g{A} \g{X}$ as follows
 \begin{equation}
\label{eq:recoveryX}
\widehat{\g{X}} = \mbox{Argmin}_{\g{X}} \quad {\left\| \widehat{\g{Y}^\circ} - \g{A}\g{X}  \right \|_F^2}.
\end{equation} 
Since $\g{A}$ is invertible, the solution of this problem is given by $\widehat{\g{X}} = \g{A}^{-1}\widehat{\g{Y}^\circ}$.\\
\end{itemize}

\subsection{Combined sparse reconstruction}

Despite its simplicity, this two-stage approach suffers from a major
drawback: the mixed components $\g{Y}^\circ$ won't be
perfectly estimated, especially when only few measurements in $\g{Y}$ are available and when noise contaminates the data. These mis-estimation errors will be amplified in the unmixing stage. Since the mixing matrix $\g{A}$ is likely to be ill-conditioned, this errors will largely impact the reconstruction accuracy of the reconstruction process.\\

A more effective strategy consists in jointly tackling both the
compressed sensing recovery and unmixing problems. Extending standard
reconstruction procedures yields the following optimization problem

\begin{equation}
\label{eq:recovery_rFISTA}
\widehat{\g{X}} = \mbox{Argmin}_{\g{X}} \quad \left
  \|\boldsymbol{\Lambda} \odot \left(\g{X} \boldsymbol{\Psi}^T\right)
\right\|_{\ell_1} + {\frac{1}{2} \left\| \g{Y} -  \g{A} \g{X}\boldsymbol{\Phi}  \right \|_F^2},
\end{equation}
where the $2\times N$ matrix $\boldsymbol{\Lambda}$ stands for the same
weight matrix we introduced in the 2-stage approach (see
section \ref{sec:param_tuning}). In the next, we make use of the reweighted
FISTA algorithm (see Appendix \ref{anex:2}) to solve
\eqref{eq:recovery_rFISTA}.  

\subsection{Constrained sparse reconstruction}

Further improving the accuracy of the components' recovery requires
imposing additional, more physical, constraints on $\g{X}$. In the
context of polarimetric data, each component $\g{x}_S$ and $\g{x}_P$ has
naturally non-negative samples. As well, under active polarized
illumination, and assuming purely depolarizing samples, the components
must verify the following inequality: $\g{x}_S \succeq \g{x}_P$. In this
section, we propose extending the reweighted FISTA algorithm to
enforce these additional constraints. The optimization problem to
tackle is described as follows

\begin{equation}
\label{eq:recovery_GFB}
\begin{multlined}
\mbox{Argmin}_{\g{X}}  \quad \left
  \|\boldsymbol{\Lambda} \odot \left(\g{X} \boldsymbol{\Psi}^T\right)
\right\|_{\ell_1} + i_{\g{X} \succeq 0}(\g{X}) \\
+ i_{\g{DX} \succeq 0}(\g{X}) + {\frac{1}{2} \left\| \g{Y} -  \g{A} \g{X}\boldsymbol{\Phi} \right \|_F^2},
\end{multlined}
\end{equation}
where $i_{\g{X} \succeq 0}(\g{X})$ stands for the characteristic function
of the positive orthant $\{ \g{X}; \; \forall i,j, \; [\g{X}]_{ij} \geq 0 \}$ and $i_{\g{DX} \succeq
  0}(\g{X})$ for the characteristic function of the convex set $\{
\g{X}; \; \forall i,j, \; [\g{DX}]_{ij} \geq 0 \}$ where $\g{D} = [1,-1]$.\\

In contrast to the standard problem in Eq.~(\ref{eq:recovery_rFISTA}),
the problem in Eq.~\eqref{eq:recovery_GFB} is composed of a sum of
convex penalizations that cannot be tackled with the FISTA
algorithm. For that end, the Generalized Forward Backward (GFB)
algorithm \cite{Raguet_13_GeneralizedForwardBackward} is the perfect
algorithm to solve such an optimization problem. The application of
the GFB to Eq.~\eqref{eq:recovery_GFB} is described in
Alg.~\ref{alg_cGFB}, where $\gamma$ stands for the gradient path
length used in the algorithm, and $L$ denotes the maximum iteration
number.

\begin{algorithm}
	\caption{Combined GFB}\label{alg_cGFB}
	\begin{algorithmic}[1]

                \State Choose $\boldsymbol{\Lambda}$ (see \ref{sec:param_tuning}), $\gamma <
                \frac{1}{\|\g{A}\|^2_2}$, $\{\mu_i\}_{i =
                  1,\cdots,3}$ such that $\sum_{i=1}^3 \mu_i = 1$,
                $\g{X}^{(0)}$, $\{\g{U}_i\}_{i=1,\cdots,3}$
                
        \While {$l<L_{outer}$}\\

		\While {$t<L$}\\

		\State $\bullet$  Gradient of the data fidelity
                term:

                $$
                \g{G}= - \g{A}^T\left(\g{Y} -\g{A} \g{X}^{(t)}\boldsymbol{\Phi}\right)\boldsymbol{\Phi}^T
                $$
                
                \State $\bullet$ Sparsity penalization
                $\g{U}_1$ :\\

                $$
                \begin{multlined}
                \g{U}_1^{(t+1)} = \g{U}_1^{(t)} + \\
                \mbox{prox}_{\gamma \|
                  \boldsymbol{\Lambda}^{(l)} \odot ( \, .\,
                  \boldsymbol{\Psi}^T)\|_{\ell_1}}\left( 2 \g{X}^{(t)}
                  - \g{U}_1^{(t)} - \gamma \g{G}  \right) - \g{X}^{(t)}
                  \end{multlined}
                  $$

                \State $\bullet$ Positivity constraint variable
                $\g{U}_2$ :\\

                $$
                \begin{multlined}
                \g{U}_2^{(t+1)} = \g{U}_2^{(t)} + \\
                \mbox{prox}_{ i_{\g{X} \geq 0}(.)}\left( 2 \g{X}^{(t)}
                  - \g{U}_2^{(t)} - \gamma \g{G}  \right) - \g{X}^{(t)}
                  \end{multlined}
                  $$

                \State $\bullet$  Components' inequality
                $\g{U}_3$ :\\

                $$
                \begin{multlined}
                \g{U}_3^{(t+1)} = \g{U}_3^{(t)} + \\
                \mbox{prox}_{ i_{\g{DX} \geq 0}(.)}\left( 2 \g{X}^{(t)}
                  - \g{U}_3^{(t)} - \gamma \g{G}  \right) - \g{X}^{(t)}
                  \end{multlined}
                  $$

		\State $\bullet$   Polarimetric components $\g{X}$:\\

                $$
                \g{X}^{(t+1)} = \sum_{i=1}^3 \mu_i \g{U}_i^{(t+1)}
                $$
                
		\EndWhile

                \State Update the weights $\boldsymbol{\Lambda}^{(l)}$ -
                see section \ref{sec:param_tuning}.\\

                \EndWhile

	\end{algorithmic}
\end{algorithm}

In this algorithm, the application of each penalization or constraint
is performed independently on distinct intermediate variables
$\{ \g{U}_i\}_{i=1,\cdots,3}$. Updating each of these variables only
requires the current estimate of the polarimetric components
$\g{X}^{(t)}$, the gradient of the quadratic data fidelity term
$\g{G} = - \g{A}^T\left(\g{Y} -\g{A}
  \g{X}^{(t)}\boldsymbol{\Phi}\right)\boldsymbol{\Phi}^T$ as well as the
so-called proximal operator of the penalization or constraint. The
proximal operators required in the above algorithms are defined in Appendix~\ref{anex:2}.\\
The convergence of the GFB algorithm is guaranteed as long as the
gradient path length verifies $\gamma < \frac{1}{\| \g{A} \|_{2}^2}$,
where the spectral norm of $\g{A}$ is defined as its largest singular
value. The scalar weights $\{ \mu_i \}_{i = 1,\cdots,3}$ must be
strictly positive and have to sum to $1$. Hence, the final update of
the polarimetric components is a weighted average of the different
intermediate variables. In the remainder of this paper, these weights
will all be set equal to $1/3$. The proposed GFB-based algorithm is
initialized with the polarimetric signals provided by the reweighted
FISTA algorithm described in the above section. Since the problem
\ref{eq:recovery_GFB} is convex, this initialization does not change
the solution but it dramatically reduces its computational cost.

\subsection{Optimization of algorithm parameters}
\label{sec:param_tuning}
The three above signal recovery approaches require tuning a certain
number of parameters, whose setting is essential for an accurate
estimation:

\begin{itemize}

\item{\bf Sparse signal representation:} the choice of the sparse
  signal representation $\boldsymbol{\Psi}$ highly depends on the
  geometrical content of the components $\g{X}$. For instance, if the
  signal $\g{X}$ is assumed to be composed of oscillatory structures,
  it is customary to choose $\boldsymbol{\Psi}$ as a discrete cosine
  transform or its localized variant. In case $\g{X}$ mainly contains
  local anisotropic contour-like features, the curvelet transform is a
  good fit. In this framework, redundant wavelets are a generic choice
  that generally provides good recovery results for most
  reconstruction problems that involve natural images. In this
  article, $\boldsymbol{\Psi}$ will be chosen as an undecimated
  wavelet frame \cite{starck:sta06}. Strictly speaking, undecimated
  wavelet frames are not orthogonal representations, which entails
  that the proximal operator defined in Eq.~\eqref{eq:proxL1} of Appendix~\ref{anex:2} is an
  approximation. Nevertheless, it is customary to use
  Eq.~\eqref{eq:proxL1} along with undecimated wavelets. Indeed, in
  that specific case, the Gram matrix of the representation
  $\boldsymbol{\Psi}^T \boldsymbol{\Psi}$ is diagonally dominant, which makes it close to an
  isometry.

\item{\bf Regularization parameters:} whether it is in the 2-step
  reconstruction approach, or in the combined reconstruction
  approaches including the reweighted FISTA or the proposed
  constrained GFB, the regularization parameters contained in matrix 
  $\boldsymbol{\Lambda}$ aim at balancing between the sparsity
  constraint and the data fidelity term. These parameters define
  thresholds that are applied to the expansion coefficients of $\g{X}$ in the sparse
  representation $\boldsymbol{\Psi}$, which eventually act as a
  denoising procedure. Therefore, in practice, these parameters are
  chosen so as to reject noise-dominated coefficients in
  $\boldsymbol{\Psi}$. For that purpose, the weight matrix is built so
  that each of its elements are the product of two terms:
  $[\boldsymbol{\Lambda}]_{ij} = \lambda_i w_{ij}$ where $\lambda_i >0
  $ and  $0 < w_{ij} \leq 1$.\\
  The first term defines the global threshold per polarimetric
  component and its value is derived straight from the derivative of
  the data fidelity term:
$$
\forall i\in\{1,2\}; \quad \lambda_i = \tau\,.\, \mbox{mad}\left( [\g{G}
  \boldsymbol{\Psi}^T ]_i \right),
$$
where the median absolute deviation ($\mbox{mad}$) is a robust empirical
estimator of a Gaussian noise standard deviation, $\bf G$ is the
gradient of the data fidelity term, and $\tau$ is a
scalar that is generally chosen between $2$ and $3$. This choice holds true for the three reconstruction approaches.\\
The extra parameters $\{w_{ij}\}$ are the standard parameters that are
defined in reweighted $\ell_1$ techniques. Following
\cite{CandesIRL108}, these parameters are chosen based on some
estimate $\widehat{\g{X}}$ of the polarimetric components
\begin{equation}
\forall i,j; \quad w_{ij} = \frac{\epsilon}{\epsilon + \frac{\left|[\widehat{\g{X}} \boldsymbol{\Psi}^T]_{ij}\right|}{\|\widehat{\g{X}} \boldsymbol{\Psi}^T\|_\infty}},
\end{equation}
where $\epsilon$ is a small scalar. This procedure is applied to the
reweighted FISTA algorithm as well as the constrained GFB.\\

\item{\bf Number of iterations, reweighted steps and stopping
    criterion : } In the next, the maximum number of iterations is
  fixed to $L=20000$, except specified otherwise. For the reweighted
  algorithms, two reweighted steps ($L_{outer}=2$) taking place respectively after
  $2000$ and $4000$ iterations were sufficient to maximize
  reconstruction quality. Lastly, for all reconstruction algorithms,
  the stopping criterion is based on the relative variation of the
  solution $\widehat{\g{X}}$ between two consecutive iterations:
  $ \|\widehat{\g{X}}^{(t+1)} - \widehat{\g{X}}^{(t)}\|_F / \|
  |\widehat{\g{X}}^{(t)}\|_F< \epsilon$, where $\epsilon = 10^{-9}$.

\end{itemize}

\section{Numerical results}\label{results}

In this section, we analyze the performance of the reconstruction
algorithms and regularization procedures described above. These
algorithms will also be compared in terms of robustness to some of the
experimental imperfections mentioned in Section~\ref{principle}. This
analysis will be conducted on a 1D test signal for the sake of
computation speed. The quality of the reconstructed signals will be
standardly evaluated throughout this section by computing the Peak
Signal to Noise Ratio (PSNR) of a concatenated
vector of the reconstructed polarimetric components, i.e.,
$[\widehat{\g{x}}_S\ \widehat{\g{x}}_P]$. Then, we will present some
polarimetric imaging numerical results on 2D signals that demonstrate
the ability of this concept of 2-pixel polarimetric camera to provide
satisfactory compressed polarization contrast images at low cost and
low complexity.

\subsection{Comparison of algorithms performance and robustness}

\subsubsection{Description of 1D test signal}

\begin{figure}[t]
\begin{center}
\includegraphics[width=8.5cm]{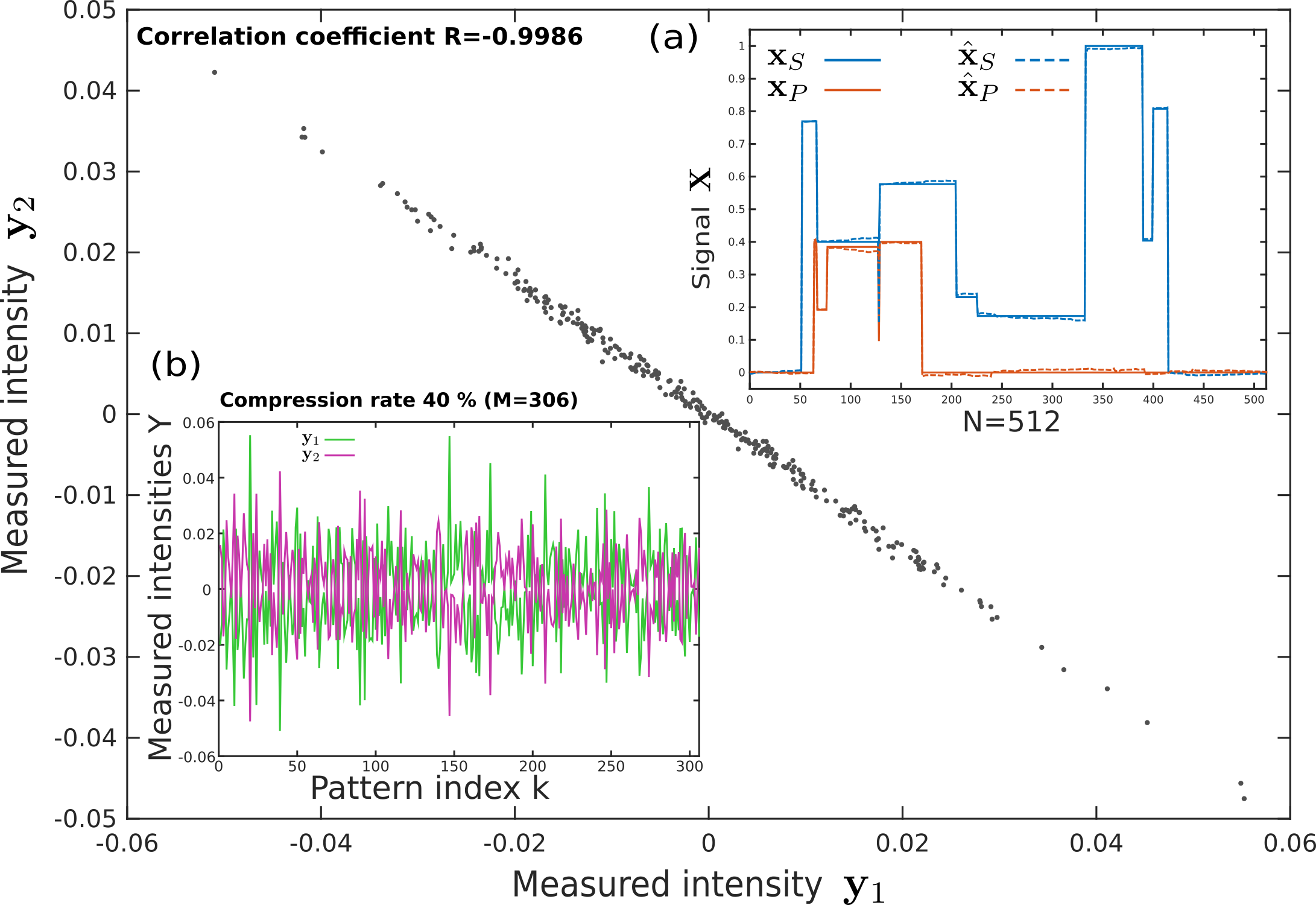}
\end{center}
\caption{Inset (a): Synthetic 1D polarimetric test signal used to
  assess recontruction algorithms performance. An example of
  reconstructed signal is also given (see text for details). Inset
  (b): example of measured intensities on photodetectors $P_1$ and
  $P_2$ for $M=306$ (compression 40 $\%$) different binary patterns
  (Hadamard) applied on the DMD. Main figure: plot of intensity
  $\g{y}_2$ as a function of $\g{y}_1$ revealing strong
  anticorrelation between the two detected
  signals. \label{fig:signals}}
\end{figure}

In order to optimize computation resources, we generated a synthetic
polarimetric 1D data sample of length $N=512$, that will be used
throughout this subsection to compare the performance of the above
algorithms. The corresponding signals $\g{x}_S$ and $\g{x}_P$ are
plotted in the inset (a) of Fig.~\ref{fig:signals}, respectively with
blue and red solid lines. The simulated polarimetric components verify
the positivity constraint $\g{x}_S \succeq \g{x}_P$, and it can be
noted that their supports are not joint, for the sake of
generality. In the inset (b) of Fig.~\ref{fig:signals}, we plot a set
of simulated detected intensity signals $\g{y}_1$ and $\g{y}_2$, with
40 $\%$ compression rate ($M=306$), and SNR = 40 dB. The mixing matrix
$\mathbf{\mathrm{A}}$ used to generate the data has been calculated as
detailed in Appendix~\ref{anex:1}, assuming an incidence angle
$\theta=50 ^\circ$ and wavelength of 780 nm (i.e., the optimal
conditions identified in Section~\ref{sec:expcond}). The patterns used
on the DMD to sample the image were generated from a randomly
scrambled Hadamard transform of the signal $\g{X}$ of size
$N=512$. Unless otherwise specified, the same experimental conditions
will be assumed for all numerical results presented in this
article. In Fig.~\ref{fig:signals}, $\g{y}_2$ is plotted as a function
of $\g{y}_1$, evidencing the strong anticorrelation existing between
the two detected signals. Lastly, in the inset (a) of
Fig.~\ref{fig:signals}, we show an example of reconstructed signals
$\hat{\g{x}}_S$ and $\hat{\g{x}}_P$ obtained with the reweighted FISTA
algorithm, and yielding PSNR~=~37.7 dB. As for all reconstructions of
the 1D signal presented below, the reconstruction process makes use of
the unidimensional undecimated wavelet transform with the Haar filter,
which is well suited to sparesely represent piecewise constant
signals. It can be checked in the inset (a) of Fig.~\ref{fig:signals}
that the algorithm has not been constrained to ensure positivity of
either $\hat{\g{x}}_S$,
$\hat{\g{x}}_P$ or $\hat{\g{x}}_S - \hat{\g{x}}_P$.\\

\subsubsection{Algorithms performance}

\begin{figure}[t]
\begin{center}
\includegraphics[width=8cm]{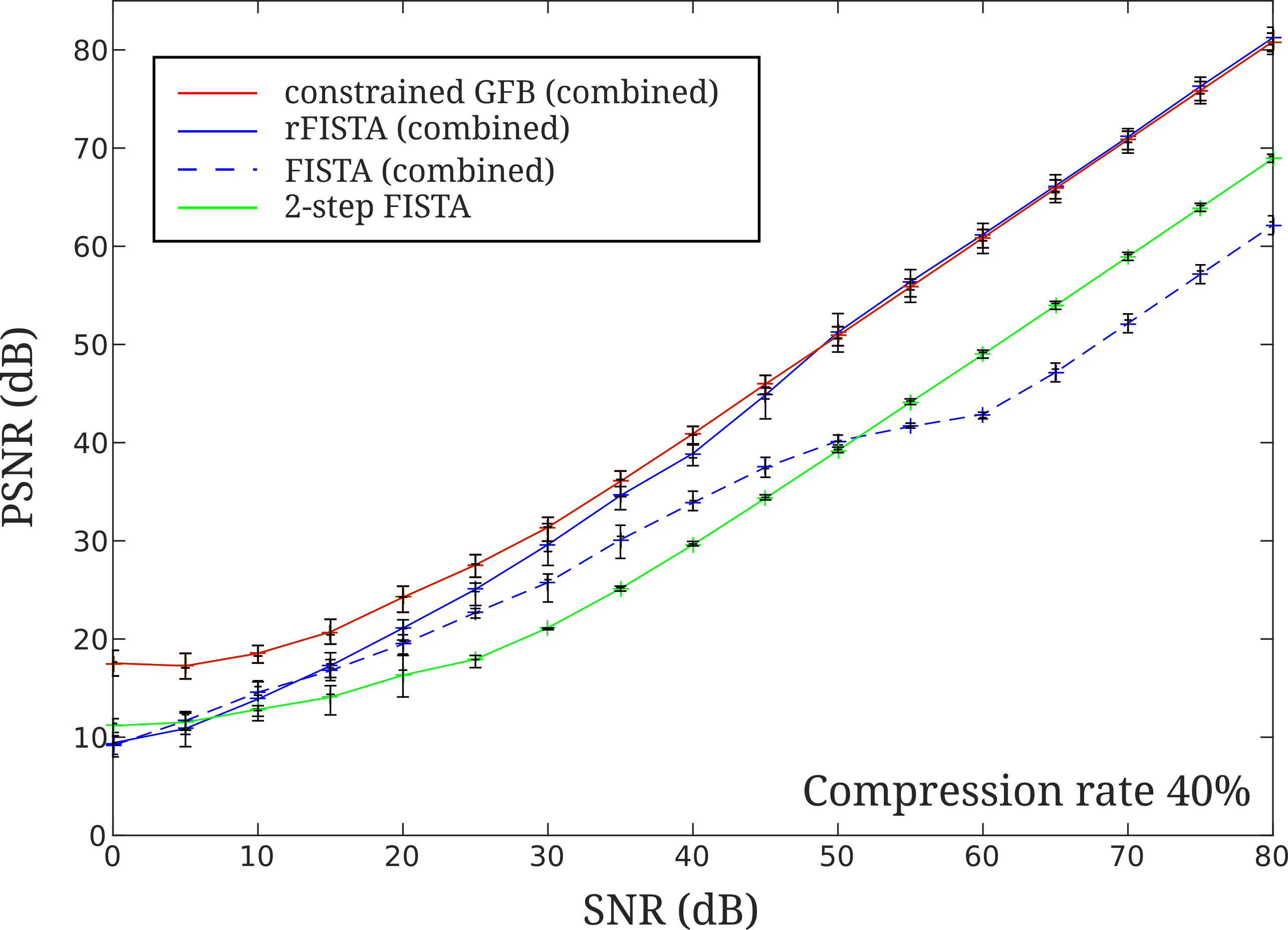}
\end{center}
\caption{Evolution of the PSNR of the signal $\hat{\g{X}}$
  reconstructed with the 4 compared algorithms as a function of
  detected signal SNR for a compression rate of 40~$\%$. The symbols
  represent the mean PSNR over 30 realizations with error bars
  indicating the interquartile range. The lines are only guides for
  the eyes.\label{fig:PSNR-SNR}}
\end{figure}

Using this 1D test signal, we are now able to compare the performance
of the various algorithms as a function of the different parameters
involved in the imaging process. Let us first analyze the influence of
the data SNR on the reconstruction quality, for an intermediate
compression rate of 40 $\%$. The evolution of the PSNR of
$\hat{\g{X}}$ as a function of the SNR is given in
Fig.~\ref{fig:PSNR-SNR}. All the data points in
Fig.~\ref{fig:PSNR-SNR} have been obtained from 30 realizations of the
numerical experiments, with the error bars indicating the
interquartile range, i.e., distance between the first to the third
quartile of the data series.

It can first be seen that all the algorithms asymptotically exhibit a
linear evolution of their PSNR as a function of the SNR. Then, it is
interesting to note that for intermediate values of SNR (10
dB$<$SNR$<$50 dB), the 2-step approach underperforms with respect to
the simplest implementation of the combined reconstruction approach
(denoted by combined-FISTA). However, as soon as a reweighted
procedure is implemented, solving the CS and the unmixing problems
simultaneously (algorithm denoted as combined-rFISTA) provides an
asymptotical gain of about 12 dB in PSNR with respect to the 2-step
algorithm. Lastly, imposing physical positivity constraints on
$\hat{\g{x}}_S$, $\hat{\g{x}}_P$ and $\hat{\g{x}}_S - \hat{\g{x}}_P$
through the implementation of the GFB algorithm (denoted as
combined-GFB) does not bring any additional gain in performance for
highest values of SNR. However, in noisy situations, for SNR~$<$~50 dB,
the positivity constraints prove efficient to improve the
reconstruction quality. A maximum gain of almost $10$~dB is obtained
for SNR~=~$0$ dB, a situation where the unconstrained rFISTA approach
fails to surpass the reconstruction quality obtained with the 2-step
procedure.

\begin{figure}[t]
\begin{center}
\includegraphics[width=8cm]{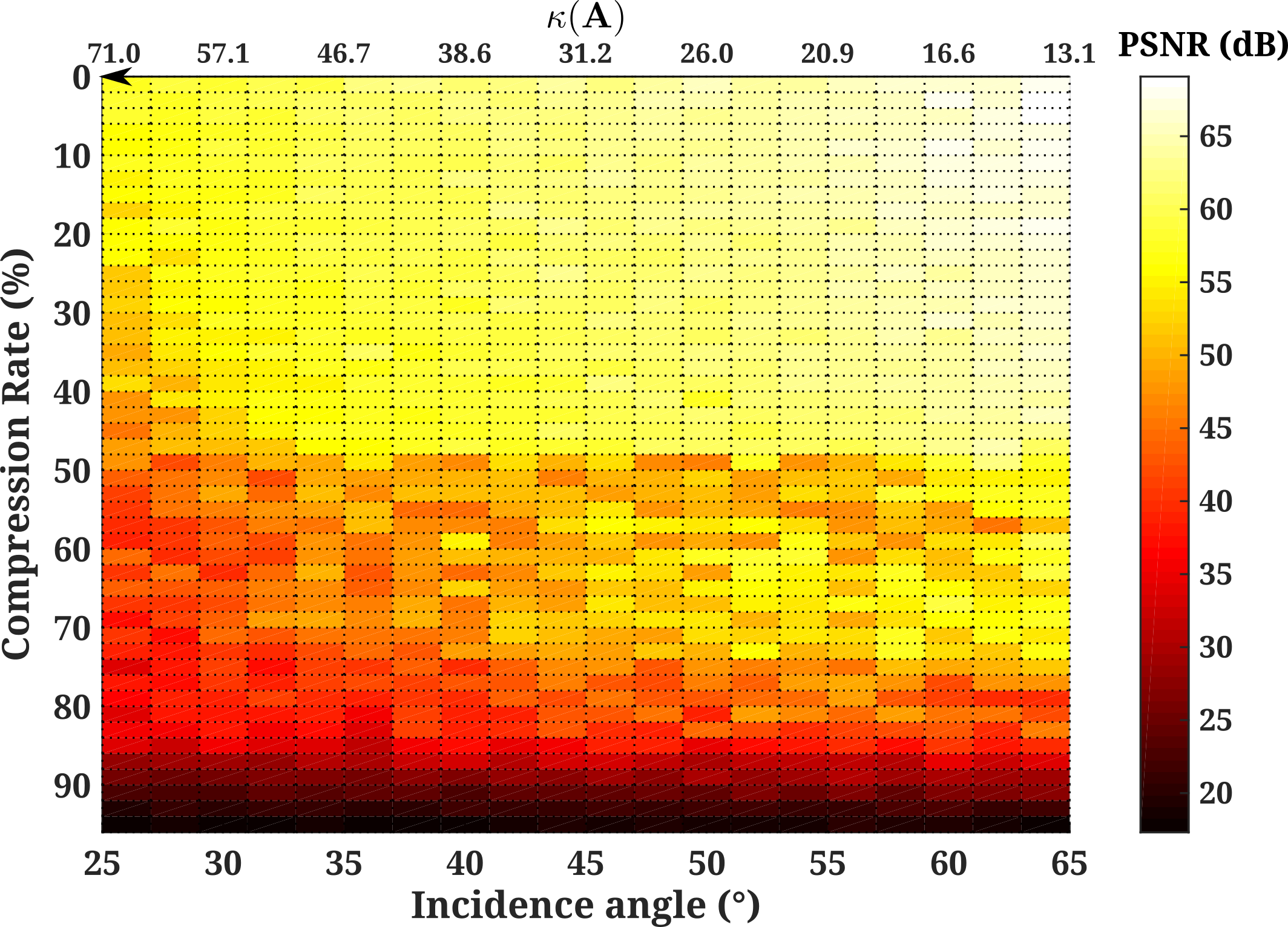}
\end{center}
\caption{2D map of the PSNR (averaged over 10 realizations) of
  $\hat{\g{X}}$ reconstructed with the combined-reweighting FISTA
  algorithm, as a function of incidence angle $\theta$, and
  compression rate. The corresponding condition number $\kappa(\g{A})$
  is indicated as a function of $\theta$. \label{fig:diag}}
\end{figure}

We now analyze the influence of the compression rate and of the
incidence angle $\theta$ on the PSNR of the reconstructed signals. As
discussed in Section~\ref{sec:expcond}, the incidence angle controls
the condition number $\kappa(\mathrm{\mathbf{A}})$ of the mixing
matrix, and hence the difficulty of the unmixing problem.  For this
numerical experiment, we consider only the combined-rFISTA algorithm
(plotted in blue solid line in the previous figure), with fixed number
of iterations ($L=10^4$), and with a SNR of 60 dB. A 2D-plot of the
average PSNR obtained over 10 realizations of each numerical
experiment is provided in Fig.~\ref{fig:diag}, for 40 values of
$\theta$ ranging between $25^\circ$ to $65^\circ$, and 49 values of
compression rate between 0 $\%$ (no compression) to 96 $\%$. It can be
observed that the reconstruction quality obeys a classical ``phase
transition'' behaviour frequenty observed in CS problems, with three
main domains which can be identified. Firstly, for compression rates
below 50 $\%$, the reconstruction is almost perfect
(PSNR~$\geq 55$~dB), whatever be the value of $\theta$. Only for
highest values of $\kappa(\mathrm{\mathbf{A}})$ (i.e., for
$\theta \leq 30^\circ$), the algorithm fails is reaching a PSNR of 50
dB, but remains above 45 dB. Then, for intermediate values of
compression rate between 50~$\%$ and 80~$\%$, the reconstruction
quality gradually decreases while remaining above $35$~dB. In that
regime, the influence of the condition number
$\kappa(\mathrm{\mathbf{A}})$ appears clearly on the reconstruction
quality. Lastly, a sharp transition occurs around 85~$\%$ compression
rate, above which no satisfactory reconstruction can be obtained
regardless of $\theta$. This 2D-map is rather encouraging towards
possible experimental implementations of the 2-pixel polarimetric CS
camera. Indeed, it shows that provided a good SNR can be ensured (here
$60$~dB), polarimetric signals can be retrieved at relatively high
compression rates ($\leq 80 \%$), and for reasonable incidence angles
on the DMD.\\

\subsubsection{Robustness to incidence angle bias and tilt errors}

Before presenting results on a polarimetric imaging scenario with 2D
test signals, we report a last numerical experiment conducted on the
1D test signal to assess the robustness of the various algorithms to
experimental imperfections, such as a bias on the incidence angle
$\theta$, and random errors on the micromirrors tilt angles.

\begin{figure}[t]
\begin{center}
\includegraphics[width=8.5cm]{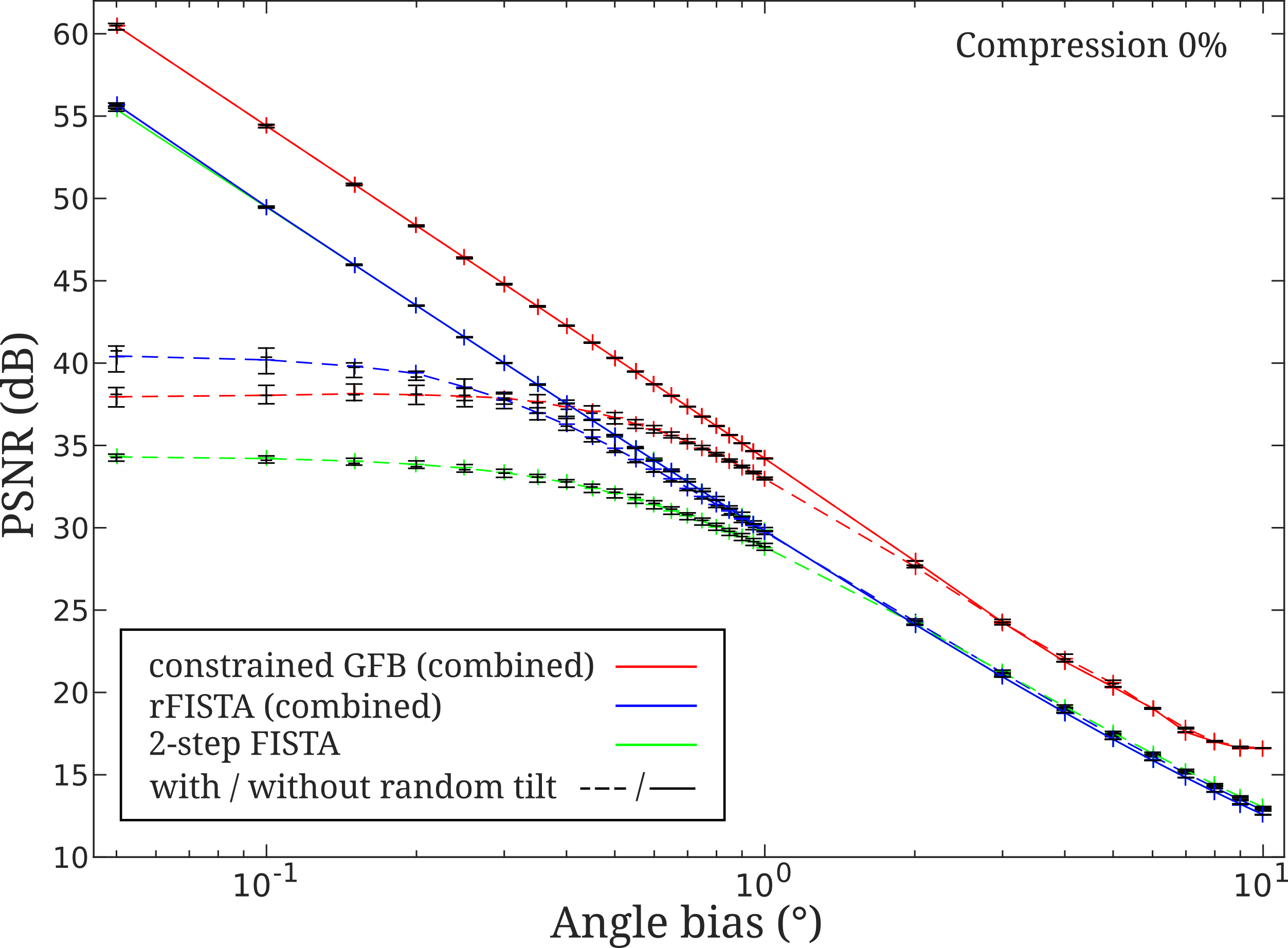}
\end{center}
\caption{Influence of the bias on incidence angle $\theta$ in the
  reconstruction quality (PSNR) for 0 $\%$ compression (solid
  lines). PSNR of the reconstructed signal in the presence of bias and
  random tilt error on individual micromirrors (dashed line). \label{fig:angle}}
\end{figure}

The influence of a bias in $\theta$ has been analyzed as
follows. Measurement vector $\g{Y}$ was generated assuming a SNR of 80
dB, and an incidence angle $\theta=50^\circ$, as in
Fig.~\ref{fig:PSNR-SNR}. However, the reconstruction procedure was run
with an incorrect mixing matrix $\g{A}$, assuming a wrong incidence
angle of $\theta+\delta\theta$. This way, we mimicked an experimental
bias $\delta\theta$ comprised between $0.05^\circ$ and $10^\circ$ on
the incidence angle. The reconstruction PSNR obtained with different
algorithms is plotted in Fig.~\ref{fig:angle} in solid lines for a
compression rate of $0$~$\%$. It can be seen that the presence of a
bias in reconstruction angle leads to a linear degradation of the PSNR
in a log-log scale for all three algorithms tested. Even for a very
small bias ($\delta \theta=0.05^\circ$), a significant drop of more
than 20 dB in reconstruction quality can be observed for all three
algorithms with respect to unbiased reconstruction, but still offering
satisfactory recovery quality (PNSR~$\geq 55$~dB). Applying physical
constraints in the reconstruction procedure with the GFB algorithm
seems to alleviate the degradation, for any magnitude of
$\delta\theta$, as far as simple angular bias is considered. However,
the PSNR value of the reconstructed vector $\hat{\g{X}}$ which is used
to gauge reconstruction quality is not satisfactory here: it has
indeed been observed on reconstructed signals that applying
constraints with an erroneous mixing matrix often leads to singular
results, where second polarimetric component $\g{x}_P$ is forced to
zero, thus yielding null polarimetric contrast. This is interpreted by
the fact that the physical constraints applied can be no longer valid
for the measured data with an incorrect matrix $\g{A}$.

Concerning the influence of random tilt errors on the micromirrors, we
simulated this effect by computing individual mixing matrices
$\mathbf{\mathrm{A}}$ for all $N$ micromirrors simulated, assuming
that the tilt angles $t_1$ and $t_2$ were affected by a random
error. For the sake of computational speed, we assumed a uniform
distribution over $11$ values of the tilt error between $\pm
1^\circ$. The PSNR of the reconstructed signal with angular bias and
uniform tilt error is also plotted in Fig.~\ref{fig:angle} in dashed
lines for 0 $\%$ compression. On the one hand, the additional random
tilts do not modify significantly the results for highest values of
angular bias ($\delta\theta \geq 1^\circ$) where the incorrect
``average'' matrix $\mathbf{\mathrm{A}}$ is responsible for most of
the quality degradation. On the other hand, for very low values of
$\delta\theta$, the PSNR reaches a limit upper value of about 35-40
dB, due to the presence of random tilt errors which seems to affect
more strongly the constrained version of the algorithm. Despite such
degradation of reconstruction quality, these numerical experiments
evidence that with imperfect experimental configurations, using a
combined recovery approach instead of a 2-step reconstruction
procedure can be advantageous. Correct PSNR values ($\geq$ 35 dB) can
be reached with the rFISTA or GFB algorithms with small angular biases
($\leq 1^\circ$) and in the presence of random tilt errors.

\subsection{Numerical polarimetric imaging results on 2D signals}

In this last section, we analyze the ability of the 2-pixel
polarimetric CS camera to retrieve polarimetric contrast images from
simulated 2D data. Owing to its performance and its simple
implementation with respect to constrained GFB, we only consider in
this section the reweighted FISTA algorithm implementing a combined
reconstruction procedure. We first consider in the next subsection a
simple imaging scenario to study the influence of polarimetric
contrast on the reconstuction quality, before a more realistic example
of polarimetric image reconstruction is given in
Section~\ref{sec:realistic}.

\subsubsection{Influence of polarimetric contrast}
\begin{figure}[t]
\begin{center}
\includegraphics[width=8.5cm]{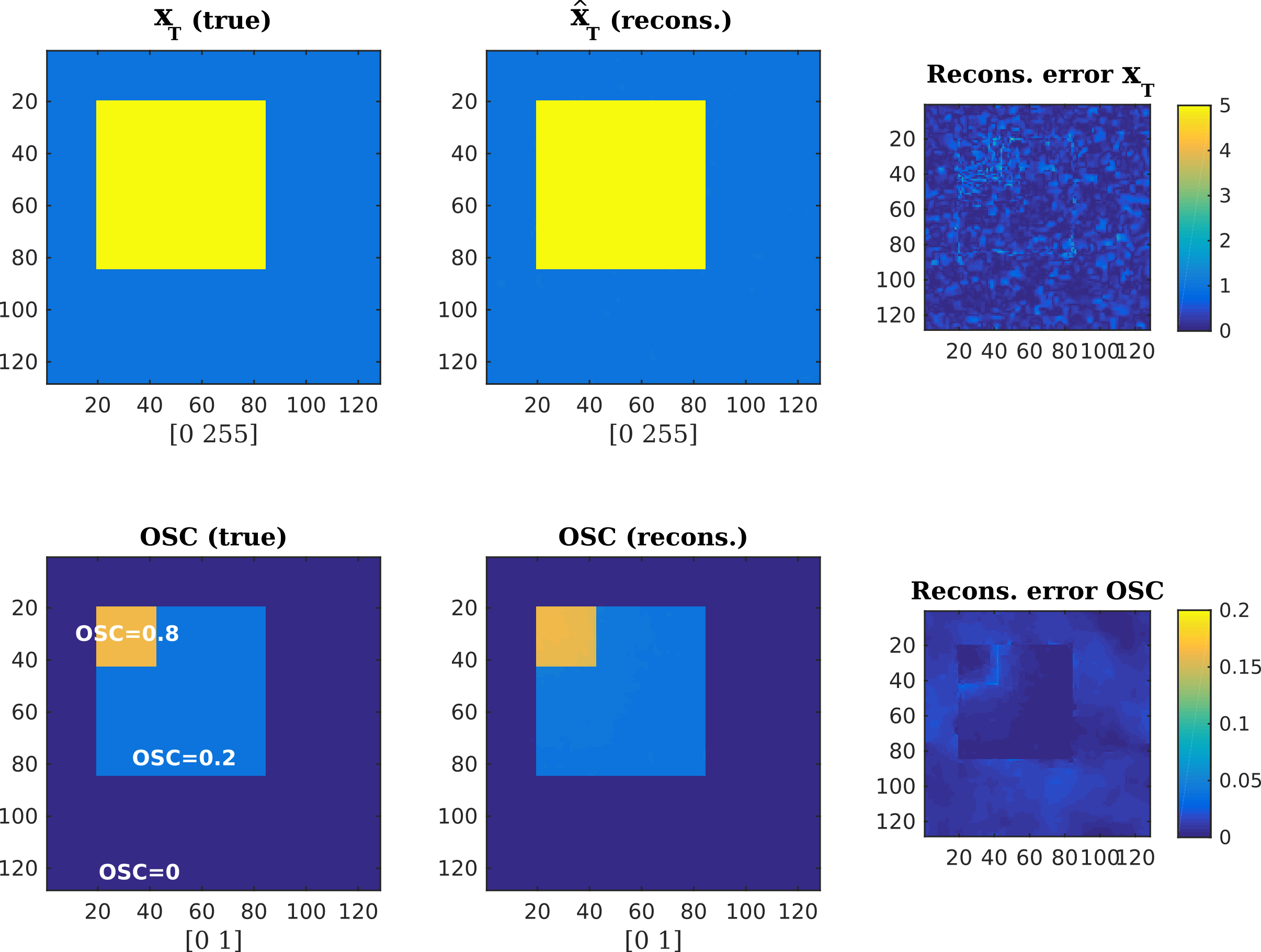}
\end{center}
\caption{Simulated polarimetric imaging scenario containing two
  objects over constant background. First row: total intensity image
  $\g{x}_T$ (true), $\hat{\g{x}}_T$(reconstruction) and reconstruction
  error map. The two objects are indistinguishable in the total
  intensity image. Second row: true and reconstructed OSC image, and
  reconstruction error map.\label{fig:square}}
\end{figure}

For this first imaging scenario, we consider a square object with
high total intensity value over a dark background, forming an image of
$N=128\times128$ pixels. This intensity image $\g{x}_T$, plotted in
Fig.~\ref{fig:square} is supposed to be the sum of two polarimetric
image components $\g{x}_S$ and $\g{x}_P$, yielding a true OSC map also
plotted in Fig.~\ref{fig:square}. In this scenario, we assume that a
first object (smaller square) cannot be distinguished from the second
object (bigger square) on the intensity image $\g{x}_T$, while OSC map
allows the two objects to be clearly identified. The background is
assumed totally depolarizing (OSC~=~$0$). The smaller square object is
always supposed slightly depolarizing (OSC~=~$0.8$) whereas the OSC of the
second object (bigger square) is varied between 0 and 1 in the
following numerical experiments. Fig.~\ref{fig:square} shows an
example of reconstruction of the total intensity and the OSC map with
the combined-rFISTA algorithm (with Haar wavelets) for SNR~=~$40$~dB,
compression rate $40\%$, incidence angle $\theta=50^\circ$ and
without bias or uncertainty on the tilt directions. The reconstruction
quality is visually very good (PSNR~=~$52.8$~dB), as evidenced by the
reconstruction error map of the total intensity and OSC shown in
Fig.~\ref{fig:square}. This simple example demonstrates the
possibility of providing relevant polarimetric information from the
proposed concept of 2-pixel polarimetric CS camera.

\begin{figure}[t]
\begin{center}
\includegraphics[width=8.5cm]{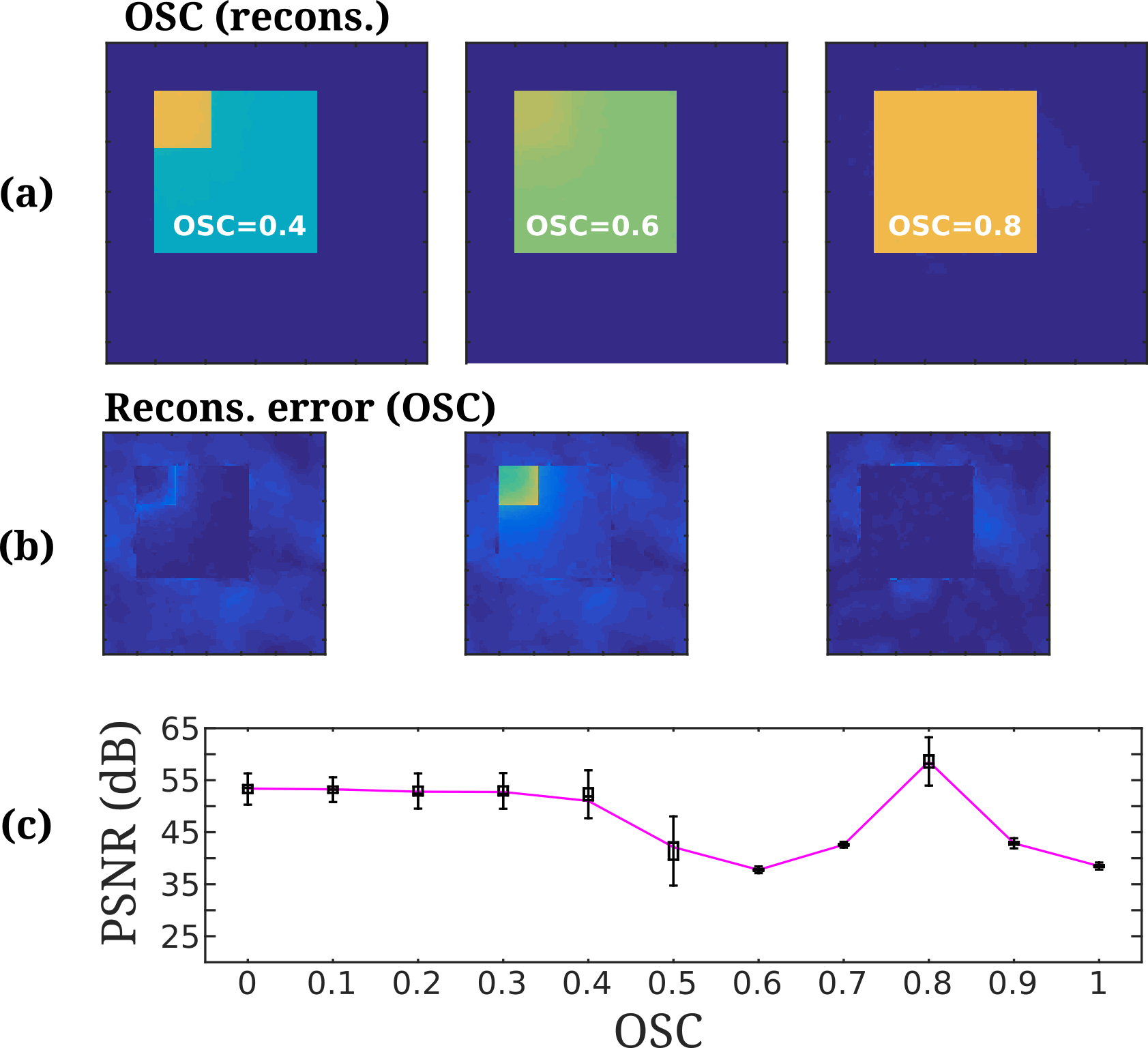}
\end{center}
\caption{(a) Reconstructed OSC image (a) and reconstruction error map
  (b) for various values of the OSC of the second object. (c)
  Evolution of the reconstruction PSNR of $\g{X}$ when the OSC of the
  second object is varied between 0 and 1. \label{fig:contrast}}
\end{figure}

In the first row (a) of Fig.~\ref{fig:contrast}, we plotted the
reconstructed OSC maps for three other values of the OSC of the second
object (bigger square). The reconstruction error map is given in the
second row (b), while the evolution of the PSNR with the OSC of the
second object is plotted in Fig.~\ref{fig:contrast}.c. These results
evidence the fact that the reconstruction quality naturally decreases
when the polarimetric contrast between the two objects is reduced,
i.e., when the reconstruction problem becomes more difficult. This can
be easily understood as the smallest variations of contrast are likely
to be burried in the noise and totally filtered out by the
regularization process. This is clearly seen for OSC~=~$0.6$, where
the sharpest features of the first object disappear in the
reconstructed OSC map. Obviously, reconstruction quality is maximimed
when the OSC of the second object reaches 0.8, i.e., a single object
is to be identified in the image, thus yielding a simpler
reconstruction problem.

\subsubsection{Example of reconstruction on realistic image data}\label{sec:realistic}

\begin{figure}[t]
\begin{center}
\includegraphics[width=8.5cm]{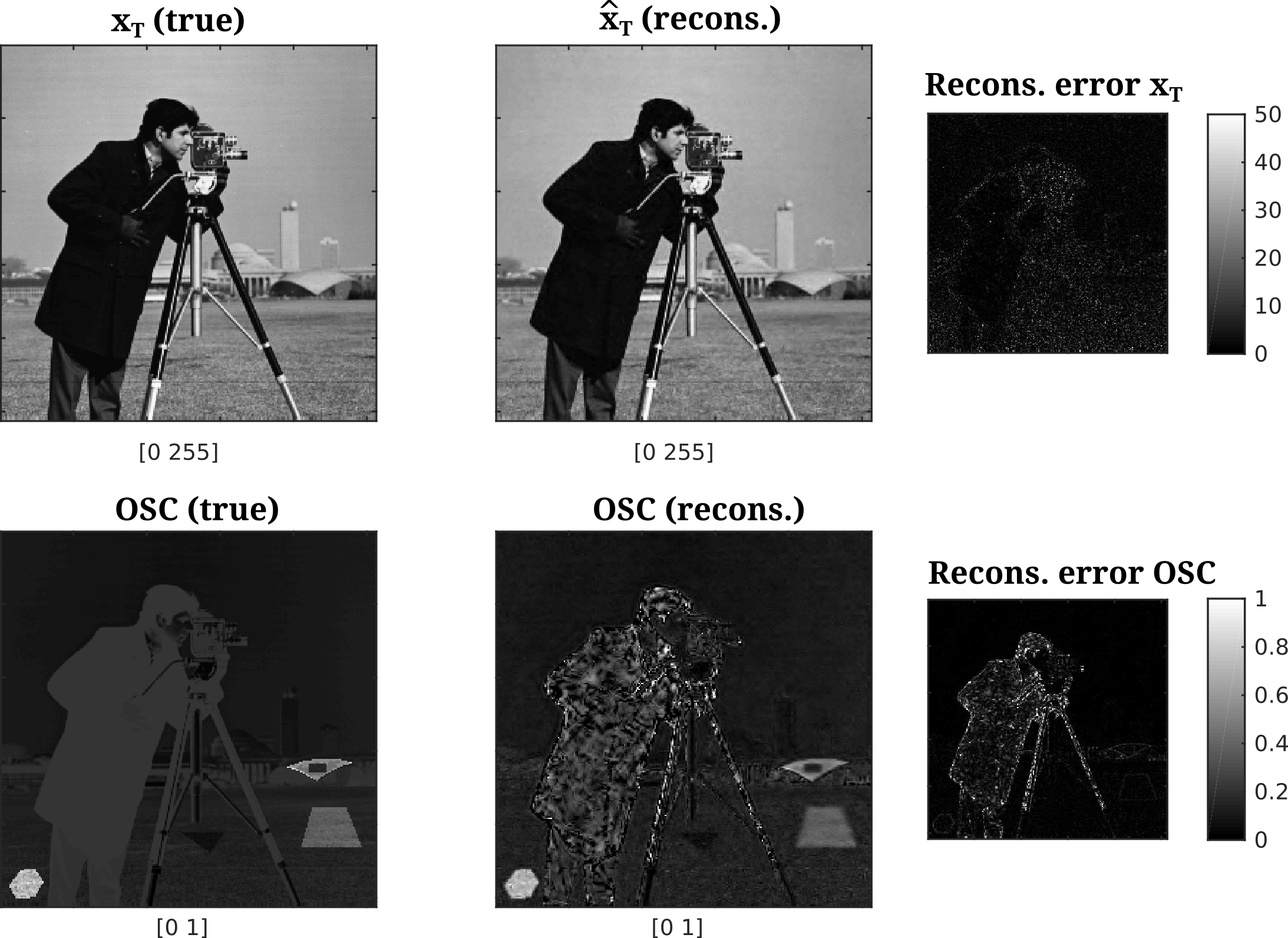}
\end{center}
\caption{Example of numerical polarimetric imaging experiment on
  realistic $512\times 512$ pixels image data. First row: total
  intensity image $\g{x}_T$, reconstruction $\hat{\g{x}}_T$ and error
  map. The total intensity image is the standard cameraman test image,
  where the 4 hidden objects are totally hidden. Second row: true OSC
  image, reconstructed OSC image and error map. The 4 hidden objects
  are revealed by their polarimetric contrast over depolarizing
  background on the OSC images. \label{fig:cam}}
\end{figure}

Laslty, we present an example of reconstructed polarimetric image on a
more realistic imaging scenario. For that purpose, we considered a
true intensity image $\g{x}_T$ of the \emph{cameraman} with size
$N=512\times 512$, as plotted in Fig.~\ref{fig:cam}. Appropriate
polarimetric components $\g{x}_S$ and $\g{x}_P$ were generated so that
a true OSC map would reveal 4 hidden objects (3 in the grass, 1 in the
buildings) over a depolarizing background, as can be seen in
Fig.~\ref{fig:cam}. The reconstruction results with Symmlet wavelet
transform are also displayed in Fig.~\ref{fig:cam}, along with
reconstruction error maps. The total intensity image is almost
perfectly reconstructed, as would be the case with a SPC imaging
system. However, the 4 hidden objects remain of course invisible in
the reconstructed image $\hat{\g{x}}_T$. Contrarily, the reconstructed
OSC map makes it possible to identify the presence of the 4 hidden
objects by revealing their polarimetric contrast over the
background. The analysis of the reconstruction error map of OSC shows
that the polarimetric information about the 4 hidden objects is fairly
retrieved. However one can notice significant reconstruction errors in
the darkest regions of the image (cameraman and tripod). These
imperfections could be lowered in the future by refining
regularization parameters and constraints in the algorithm
implementation.

\section{Conclusion and perspectives}\label{conclu}

In this article, we have proposed a new concept of CS polarimetric
imaging inspired from the SPC principle. Relying on the tiny
differences in reflection coefficients of mirrors with incidence angle
and polarization direction, the setup proposed allows intensity and
polarimetric contrast informations to be recovered from the temporal
acquisition on two single-pixel detectors, and without requiring any
polarization analysis optical component. We have shown that this
recovery problem could be analyzed as a joint CS and source separation
tasks, that can be tackled independantly and successively using
standard approaches (respectively $\ell_1$ minimization and direct
matrix inversion), or optimally treated in an original combined
reconstruction approach. For that purpose, we have presented different
versions of a combined algorithm, including FISTA implementation of
the combined optimization problem, with potential reweighted steps
that have proved efficient to increase the reconstruction
quality. Moreover, to enforce additional physical constraints on the
measured data, we have proposed a constrained sparse reconstruction
method based on the GFB algorithm, allowing the reconstruction quality
to be improved in low SNR conditions.

Numerical simulations have permitted to validate these approaches and
to analyze the influence of experimental conditions, such as incidence
angle, SNR, micromirrors tilt errors, etc. Generally speaking, these
results are encouraging towards the experimental implementation of
such an imaging setup as good reconstruction quality was obtained for
reasonable incidence angles on the mirror, even in the presence of a
small bias or randomness in the mirrors tilt direction. Significant
compression rates could be achieved while still offering sufficient
reconstruction quality, as illustrated on the 2D simulation results
presented.

Experimental implementation of this computational imaging approach appears as a natural perspective
to this work. Another interesting research track is the design of a
blind calibration method so as to compensate possible mis-estimation
of the mixing matrix $\g{A}$ involved in the reconstruction
process. More generally, applying CS concepts to more sophisticated
multi-channel polarimetric imaging techniques which are characterized
by very specific algebraic constraints is likely to raise interesting
reconstruction issues.

\renewcommand\thesection{\Alph{section}}
\setcounter{section}{0}

\section*{Appendices}

\section{Computation of the reflection coefficients}\label{anex:1}
We consider a mirror with complex refraction index
$\tilde{n}=n + j k$, which can be conveniently rewritten
$\tilde{n}=n(1+j \kappa)$ \cite{bor99}. We consider the reflection of a
beam propagating in an input medium consisting of air ($n_a=1$). For an incidence angle $\theta $
on the mirror, one has from
the generalized Snell's law $n_a \sin \theta = \tilde{n} \sin \tau$, with $\tau$
the refraction angle. Setting $u=n \cos r$ and $v=n\kappa \cos r$, one
has $\tilde{n} \cos r = u + j v$.

It can be shown \cite{bor99} that with such notations the reflection
coefficient in intensity for TE waves (S-polarized) can be written
\cite{bor99}
$$r^S(\theta)=\frac{(n_a\cos \theta - u)^2+v^2}{(n_a\cos \theta + u)^2+v^2}=\frac{(\cos \theta - u)^2+v^2}{(\cos \theta + u)^2+v^2},$$
whereas for the TM (P-polarized) waves, one has
\begin{equation}
  r^P(\theta)=\frac{(n^2(1-\kappa^2)\cos \theta -  u)^2+(2n^2\kappa \cos \theta - v)^2}{(n^2(1-\kappa^2)\cos \theta + u)^2+(2n^2\kappa \cos \theta + v)^2},
\end{equation}
with $u= \sqrt{({\cal A}+\sqrt{\cal B})/2}$ and $v= \sqrt{(-{\cal A}+\sqrt{{\cal B}})/2}$, and
${\cal A}= n^2(1-\kappa^2) - \sin^2 \theta$ and ${\cal B}= {\cal A}^2 + 4 n^4\kappa^2$.

From these equations, we were able to compute the micromirrors
reflection coefficients for any incidence and any wavelength. The
complex refraction index was obtained from a recent
evaluation of aluminum reflection coefficients in Reference
\cite{rak98}.

\section{Reconstruction algorithms}\label{anex:2}
\subsection{Forward-backward splitting algorithm}\label{anex:2_1}
Whether it is in the 2-step or in the combined sparse reconstruction
approach, recovering the polarimetric signal $\bf X$ requires solving
optimization problems of the form
$$
\min_{\bf X} \quad \mathcal{P}({\bf X}) + \mathcal{L}({\bf X}),
$$
where the data fidelity term $\mathcal{L}({\bf X})$ is a quadratic
norm that is differentiable and whose gradient is Lipschitz with some
constant $L_i$, and
$\mathcal{P}({\bf X})$ is a convex but non-smooth
penalization. Minimization problem can be carried out efficiently with
recent proximal algorithms \cite{Boyd_Proximal14}, and more precisely
with the Forward-Backward Splitting (FBS) algorithm
\cite{CombettesWajs05}. The FBS algorithm is an iterative procedure that can be
described with the following update rule at iteration $t$
$$
{\bf X}^{(t+1)} = \mbox{prox}_{\gamma \mathcal{P}}\left({\bf X}^{(t)}
  - \gamma \nabla \mathcal{L}({\bf X}^{(t)}) \right)
$$
where $\nabla \mathcal{L}({\bf X}^{(t)})$ stands for the derivative of
$\mathcal{L}$ in ${\bf X}^{(t)}$ and $\gamma$ is the gradient path
length. The proximal operator $\mbox{prox}_{\gamma \mathcal{P}}$ is
defined by the solution to the problem
$$
\mbox{prox}_{\gamma \mathcal{P}}({\bf Z}) = \mbox{Argmin}_{\bf X}
\gamma\mathcal{P}({\bf X}) + \frac{1}{2} \|{\bf Z} - {\bf X}\|_F^2.
$$
While the proximal operator of the penalization function is defined as
the solution of an optimization problem, standardly used proximal
operators admit a closed form expression (see Appendix~\ref{anex:2_2}).\\
In this article, the term $\mathcal{P}$ will penalize non sparse
solutions and will be based on the $\ell_1$ norm. Introduced in
\cite{CandesIRL108}, the re-weighted $\ell_1$ norm further introduces
weights $\boldsymbol{\Lambda}$ that aim at reducing the bias induced
by the standard $\ell_1$ norm. Consequently, the penalization term
used in this article will take the generic form
$$
\mathcal{P}({\bf X}) = \left\|\boldsymbol{\Lambda} \odot \left({\bf X}
  \boldsymbol{\Psi}^T\right) \right \|_{\ell_1}
$$
where $\boldsymbol{\Psi}$ is the signal representation where sparsity
is modeled. The final optimization procedure then alternates between
updates of $\bf X$ for fixed weights $\boldsymbol{\Lambda}$ and
updates of these weights, which has been showed to dramatically
improves the accuracy of the reconstruction.\\
In the proposed reweighted algorithm, updates of the polarimetric
signals are carried out using an accelerated version of the FBS
algorithm coined FISTA \cite{BeckFista09}. The generic description of
the reweighted FISTA
algorithm is given in Algorithm \ref{alg_rFISTA}

\begin{algorithm}
\caption{Reweighted FISTA  algorithm}\label{alg_rFISTA}
	\begin{algorithmic}[1]
          \State Choose initial point ${\bf X}^{(0)}$ and set the
          weights to $1$.\\
		\While {$l<L_{outer}$}\\
                \State Update the polarimetric components with
                starting point ${\bf
                 W}^{(0)} = {\bf X}^{(l-1)}$ and ${\bf
                  Z}^{(0)} = {\bf X}^{(l-1)}$. Fix $\rho^{(0)} = 1$.\\
                \While {$t<L$}\\
		${\bf W}^{(t+1)} = \mbox{prox}_{\gamma
                  \left\|\boldsymbol{\Lambda} \odot \left(\; . \;
  \boldsymbol{\Psi}^T\right) \right \|_{\ell_1}}\left({\bf Z}^{(t)}
  - \gamma \nabla \mathcal{L}({\bf Z}^{(t)}) \right)$\\
$\rho^{(t+1)} = \frac{1 + \sqrt{1 + 4 {\rho^{(t)}}^2}}{2}$\\
${\bf Z}^{(t)} = {\bf W}^{(t+1)} + \frac{\rho^{(t)} - 1}{\rho^{(t+1)}}
({\bf W}^{(t+1)}  - {\bf W}^{(t)} )$\\
\EndWhile
Fix ${\bf X}^{(l)} =  {\bf Z}^{(L)} $
		\State Update the weights $\boldsymbol{\Lambda}^{(l)}$ -
                see Section \ref{sec:param_tuning}.
		\EndWhile
	\end{algorithmic}
\end{algorithm}
This procedure generally increases the accuracy of the reconstruction
process with few updates of the weights $\boldsymbol{\Lambda}$,
typically $L_{outer}$ between $3$ to $5$. Details about practical
parameter tuning are given in Section~\ref{sec:param_tuning}.

\subsection{Useful proximal operators}\label{anex:2_2}

Hereafter, we described different proximal operators that are used in
the proposed reconstruction algorithms.

\subsubsection*{Reweighted $\ell_1$} 
Assuming that the sparse signal
representation $\boldsymbol{\Psi}$ is an orthogonal matrix, the
proximal operator of $\left
  \|\boldsymbol{\Lambda} \odot \left(\g{X} \boldsymbol{\Psi}^T\right)
\right\|_{\ell_1} $ is defined as
\begin{equation}
\label{eq:proxL1}
\mbox{prox}_{\|\boldsymbol{\Lambda} \odot ( \,
  .\,\boldsymbol{\Psi}^T)\|_{\ell_1}} \left( \g{Z}\right) =
  \boldsymbol{\Psi} \mathcal{S}_{\boldsymbol{\Lambda}} \left(  \g{Z} \boldsymbol{\Psi}^T \right)
\end{equation} 
where the weighted soft-thresholding operator
$\mathcal{S}_{\boldsymbol{\Lambda}}$ is defined as
\begin{equation}
\begin{multlined}
\forall i,j; \quad \mathcal{S}_{\boldsymbol{\Lambda}}\left(
  \g{Z}_{ij}\right) = \\
  \left\{ 
\begin{array}{l}
 \g{Z}_{ij} -\boldsymbol{\Lambda}_{ij} \mbox{sign}(\g{Z}_{ij}) \mbox{ if } |\g{Z}_{ij}| > \boldsymbol{\Lambda}_{ij}\\
 0  \mbox{ otherwise}
\end{array}
\right.
\end{multlined}
\end{equation}

\subsubsection*{Positivity constraint} 
The proximal operator of the
  positivity constraint is defined as the orthogonal projector onto
  the non-negative orthant:
\begin{equation}
\label{eq:proxPositivity}
\begin{multlined}
\forall i,j; \quad \mbox{prox}_{ i_{\g{X} \succeq 0}(.)}\left(
  \g{Z}_{ij}\right) = \\
  \left\{ 
\begin{array}{l}
 \g{Z}_{ij}  \mbox{ if } \g{Z}_{ij} > 0\\
 0  \mbox{ otherwise}
\end{array}
\right.
\end{multlined}
\end{equation}

\subsubsection*{Inequality constraint} 
The inequality constraint $x_s \succeq
  x_p$ can be recast as $\g{D} \g{X} \succeq 0$, where $\g{D} = [1 ,
  -1]$. Its proximal operator is defined as the orthogonal projector
  onto the convex set $\left \{ \g{Z}; \; \forall i,j, \; [\g{D}
    \g{Z}]_{ij} \geq 0 \right\}$, which is defined as
$$
\mbox{prox}_{ i_{\g{D}\; . \; \succeq 0}}(\g{Z}) =
\mbox{Argmin}_{\g{D}\g{X} \succeq 0} \quad \frac{1}{2}\|\g{Z} - \g{X}\|_F^2.
$$
This expression can be more conveniently recast in a Lagragian
formulation by introducing the Lagrange multipliers $\boldsymbol{\pi}$
$$
\mbox{prox}_{ i_{\g{D}\; . \; \succeq 0}}(\g{Z}) =
\mbox{Argmin}_{\g{X}} \quad <\g{D}\g{X},\boldsymbol{\pi}>+ \frac{1}{2}\|\g{Z} - \g{X}\|_F^2.
$$
Its optimum is obtained for $\g{X} = \g{Z} - \g{D}^T
\boldsymbol{\pi}$, which should verify the constraint $\g{D} \g{X}
\succeq 0$. This entails
$$
\g{D} \g{Z} - 2 \boldsymbol{\pi}  \succeq 0.
$$
Consequently, the Lagrange multipliers must take the values
\begin{equation}
\label{eq:proxPositivity_Lagrange}
\forall j; \quad \boldsymbol{\pi}_j = \left\{ 
\begin{array}{l}
 \frac{1}{2} [\g{D}\g{Z}]_j \mbox{ if } [\g{D}\g{Z}]_j < 0\\
 0  \mbox{ otherwise}
\end{array}
\right.
\end{equation}
From this expression of the Lagrange multipliers, the proximal operator
is then defined as
\begin{equation}
\label{eq:proxIneq}
\mbox{prox}_{ i_{\g{D}\; . \; \succeq 0}}(\g{Z}) = \g{Z} - \g{D}^T \boldsymbol{\pi}.
\end{equation}

\section*{Acknowledgments}
The authors would like to thank Anthony Carr\'e for his help in the 3D
rendering of Fig.~\ref{fig:setup}. This work is supported by the
European Research Council through the grant LENA (contract no. 678282).






%

\end{document}